\documentclass[runningheads]{llncs}
\usepackage{graphicx}
\usepackage{subcaption} % subfigure
\usepackage{multirow} % multirow
\usepackage{multicol} % multicol
\usepackage{amsmath}
\usepackage{amsfonts}
\usepackage{bm}
\usepackage{color} % temporary for highlighting
\usepackage{soul} % temporary for highlighting
\begin{document}
\title{Isotropic Representation Can Improve Dense Retrieval}
% \thanks{Supported by organization x.}
% \author{Anonymous authors}
\author{Euna Jung\inst{1}\orcidID{0000-0002-8148-8735} \and
Jungwon Park\inst{1}\orcidID{0000-0001-9214-3073} \and \\
Jaekeol Choi\inst{1,2}\orcidID{0000-0003-1992-0731} \and
Sungyoon Kim\inst{1}\orcidID{0000-0003-0287-5276} \and \\
Wonjong Rhee\inst{1,3}\orcidID{0000-0002-2590-8774}\thanks{Corresponding Author.}
}
% %
\authorrunning{Euna Jung. et al.}
\institute{GSCST at Seoul National University, Seoul, South Korea
\email{\{xlpczv,quoded97,jaekeol.choi,clifter0122,wrhee\}@snu.ac.kr}\\ \and
Naver Corp. \and IPAI, AIIS at Seoul National University}
% % \email{\{abc,lncs\}@uni-heidelberg.de}
% }
%
\maketitle              % typeset the header of the contribution
\begin{abstract}
The latest Dense Retrieval~(DR) models typically encode queries and documents using BERT and subsequently apply a cosine similarity-based scoring to determine the relevance. BERT representations, however, are known to follow an anisotropic distribution of a narrow cone shape and such an anisotropic distribution can be undesirable for relevance estimation. In this work, we first show that BERT representations in DR also follow an anisotropic distribution. We adopt unsupervised post-processing methods of Normalizing Flow and whitening to cope with the problem, and develop a token-wise method in addition to the sequence-wise method. We show that the proposed methods can effectively enhance the isotropy of representations, thereby improving the performance of DR models such as ColBERT and RepBERT. To examine the potential of isotropic representation for improving the robustness of DR models, we investigate out-of-distribution tasks where the test dataset differs from the training dataset. The results show that isotropic representation can certainly achieve a generally improved performance.\footnote{The code is available at \url{https://github.com/SNU-DRL/IsotropicIR.git}}
\keywords{Dense Retrieval \and Isotropic Representation \and Normalizing Flow \and Whitening \and Robustness} %\and whitening }
\end{abstract}

\section{Introduction}
\vspace{-0.2cm}

\begin{figure}
  \hspace*{\fill}%
  \subcaptionbox{Isotropy}{\includegraphics[width=0.33\linewidth]{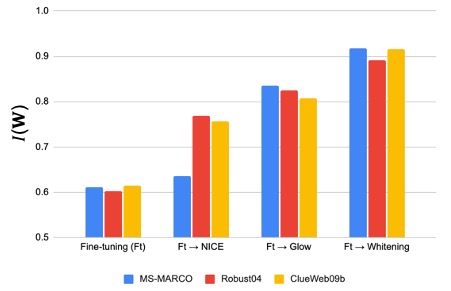}}
  \hfill
  \subcaptionbox{Isotropy and ID performance}{\includegraphics[width=0.28\linewidth]{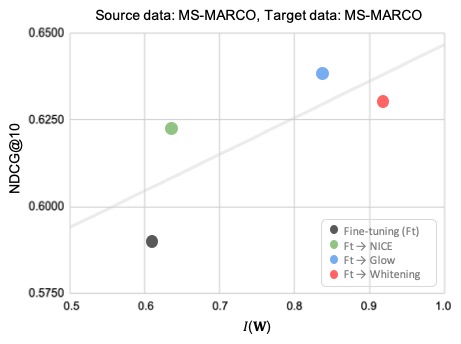}}
  \hfill
  \subcaptionbox{Isotropy and OOD performance}{\includegraphics[width=0.28\linewidth]{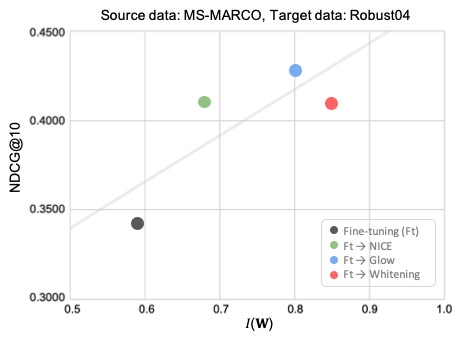}}
  \hspace*{\fill}%
  \vspace{-0.2cm}
  \caption{Isotropy and re-ranking performance of ColBERT. For increasing isotropy, three post-processing methods are studied. In (a), a metric of isotropy ($I(\mathbf{W})$) is shown for BERT representations of three dense retrieval datasets. (b) and (c) show isotropy and re-ranking performance of ColBERT on MS-MARCO for In-Distribution (ID) setting and Out-Of-Distribution (OOD) setting, respectively.}
  \label{fig:figure1}
  \vspace{-0.4cm}
\end{figure}

Recently, many Dense Retrieval (DR) models encode representations of queries and documents using BERT and estimate the relevance scores based on the simple similarity function such as cosine similarity or dot product. The representations of language models such as BERT, however, are known to follow an anisotropic distribution \cite{gao2019representation,wang2019improving,luo2020positional,kovaleva2021bert,timkey2021all,yu2022rare}. Anisotropic distribution refers to a directionally non-uniform distribution, such as a narrow cone \cite{li2020sentence}.
If representations are anisotropically distributed, relevance estimation of DR models can be misleading. Because DR employs simple similarity functions for efficient computation, it is important to alleviate this anisotropy problem.
In this study, we aim to show that BERT representations of DR also follow an anisotropic distribution and to improve the performance of BERT-based DR by enforcing isotropy to the representations.

To enhance the isotropy of DR representations, we adopt post-processing methods used in the field of sentence embedding such as Normalizing Flow~\cite{li2020sentence} and whitening~\cite{su2021whitening} and apply them on two representative DR models, ColBERT~\cite{khattab2020colbert} and RepBERT~\cite{zhan2020repbert}.
Since the post-processing methods used for sentence embedding transform the representation of each sequence (i.e. sentence), they cannot be directly applied to multi-vector DR models, for example ColBERT, that compute cosine similarity among the token representations. Therefore, we consider a token-wise transformation of representations in addition to the sequence-wise transformation. As we will show later, the token-wise transformation turns out to be useful even for RepBERT that is a single-vector DR model. We empirically show the effectiveness of post-processing methods and compare the token-wise and the sequence-wise transformations when applicable.

By enforcing isotropy to the BERT representations, we show that we can significantly improve the re-ranking performance of both ColBERT and RepBERT. Adopting Normalizing Flow or whitening increases the performance of ColBERT by 5.2\% $\sim$ 8.1\% and the performance of RepBERT by 8.5\% $\sim$ 23.3\% on NDCG at 10 (NDCG@10) across three datasets. In the experiment of RepBERT, we have found that either token-wise method or sequence-wise method can perform better depending on the characteristics of the dataset.

To examine the potential of isotropic representation beyond the basic re-ranking task in the In-Distribution (ID) setting where the source data used for training and the target data used for the test are the same, we additionally investigate Out-Of-Distribution (OOD) setting where the source data is different from the target data. We evaluate the robustness of DR models for OOD tasks following Wu et al.~\cite{wu2021neural}. With our experiments, we have found that enforcing isotropy on BERT representations can improve the robustness of DR by 5.0\% $\sim$ 25.0\% for NDCG@10. OOD performance of ColBERT trained on MS-MARCO can even surpass the ID performance of Robust04 and ClueWeb09b when the post-processing methods are applied.

We summarize our contributions in Figure~\ref{fig:figure1}. In this paper, we focus on improving both ID and OOD performances of DR models by enforcing representation isotropy.

\section{Related Works}
\vspace{-0.2cm}

\subsection{Dense Retrieval and Similarity Function}
Depending on whether DR uses a single vector or multiple vectors for encoding each of queries and documents, DR models are divided into \textit{single-vector} and \textit{multi-vector} models.
RepBERT~\cite{zhan2020repbert}, ANCE~\cite{xiong2020approximate}, and RocketQA~\cite{qu2020rocketqa} are examples of single-vector DR models. 
ColBERT~\cite{khattab2020colbert} and COIL~\cite{gao2021coil} are examples of multi-vector DR models, and they are generally known to perform better.
Both types of DR models estimate the relevance score using a similarity function between the representation vectors of query and document. For efficient computation, the similarity function needs to be decomposable such that the representations of the documents can be pre-computed and stored~\cite{karpukhin2020dense}, and an efficient Approximate Nearest Neighbor~(ANN) retrieval~\cite{johnson2019billion} can be performed.
Most of the decomposable similarity functions are based on cosine similarity~\cite{mussmann2016learning}.

In this paper, we focus on two representative DR models, RepBERT and ColBERT\footnote{To clearly compare RepBERT and ColBERT, we simplify ColBERT to use the same encoders for both queries and documents and skipped the final projection layer that reduces the representation dimension. Also, we use cosine similarity instead of the dot product for RepBERT.}. RepBERT encodes \textit{sequence representation} for each query and document by summing up token representations and estimates the relevance score using the dot product between the \textit{sequence representations}. On the other hand, ColBERT estimates the relevance score by summing up the cosine similarity values between \textit{token representations}.
\vspace{-0.2cm}

\subsection{Anisotropic Distribution of BERT Representations}
\label{sec:2.2}
Representations of large-scale language models are known to follow anisotropic distributions \cite{gao2019representation,wang2019improving,luo2020positional,kovaleva2021bert,timkey2021all,yu2022rare}.
For example, some studies~\cite{luo2020positional,kovaleva2021bert,timkey2021all} show the existence of outlier dimensions having extreme values in representation vectors and attribute the anisotropic distribution to the outlier dimensions.
As this anisotropic distribution can negatively affect the performance of sentence embeddings, Li et al.~\cite{li2020sentence} and Su et al.~\cite{su2021whitening} applied Normalizing Flow and whitening respectively for the sentence embedding, where the cosine similarity between two sentences' BERT representations is used for the relevance estimation.
DR is different from sentence embedding in that query and document have distinct characteristics, and some multi-vector DR models utilize token-wise similarity. In this DR study, we show that the representations of BERT-based DR models also suffer from the anisotropic distribution and consider token-wise transformation as well as sequence-wise transformation.
\vspace{-0.2cm}

\subsection{Robustness of Ranking Models}
\label{subsec:robustness}
The robustness of ranking models refers to the ability of models to operate properly in abnormal situations, which is an essential factor for ranking models in the real world~\cite{wu2021neural}.
While the robustness of the ranking models can be defined with multi-dimensional factors~\cite{goren2018ranking}, OOD generalizability can be regarded as one of the most important factors for NRMs, which tend to show relatively poor performance for OOD tasks compared to the traditional ranking models~\cite{wu2021neural}.
Some of the existing works have focused on improving OOD generalizability of ranking models~\cite{thakur2022domain,wu2021neural}.
However, it was difficult to show a case where the OOD performance surpasses the ID performance.
In this study, we show that the OOD performance of NRMs can be sufficiently improved by enforcing representation isotropy such that OOD outperforms the baseline ID.

\section{Methodology}
\vspace{-0.2cm}

\subsection{Enforcing Isotropy} 
\label{subsection:enforcing_isotropy}
The baseline method of training a BERT-based DR model is shown in Figure~\ref{fig: methods}(a). In our work, we enhance the isotropy of BERT representations with Normalizing Flow or whitening as shown in Figure~\ref{fig: methods}(b) and \ref{fig: methods}(c). The two post-processing processes follow entirely unsupervised frameworks.

\begin{figure*}[t]
  \setcounter{figure}{1}
  \centering
  \begin{subfigure}[b]{0.18\textwidth}
    \centering
    \includegraphics[width=\linewidth]{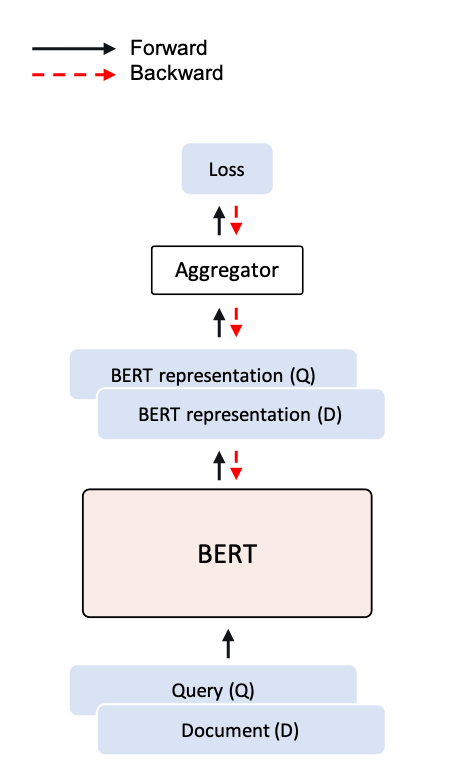}
    \caption{Fine-tuning}
  \end{subfigure} \hfill
  \begin{subfigure}[b]{0.38\textwidth}
    \centering
    \includegraphics[width=\linewidth]{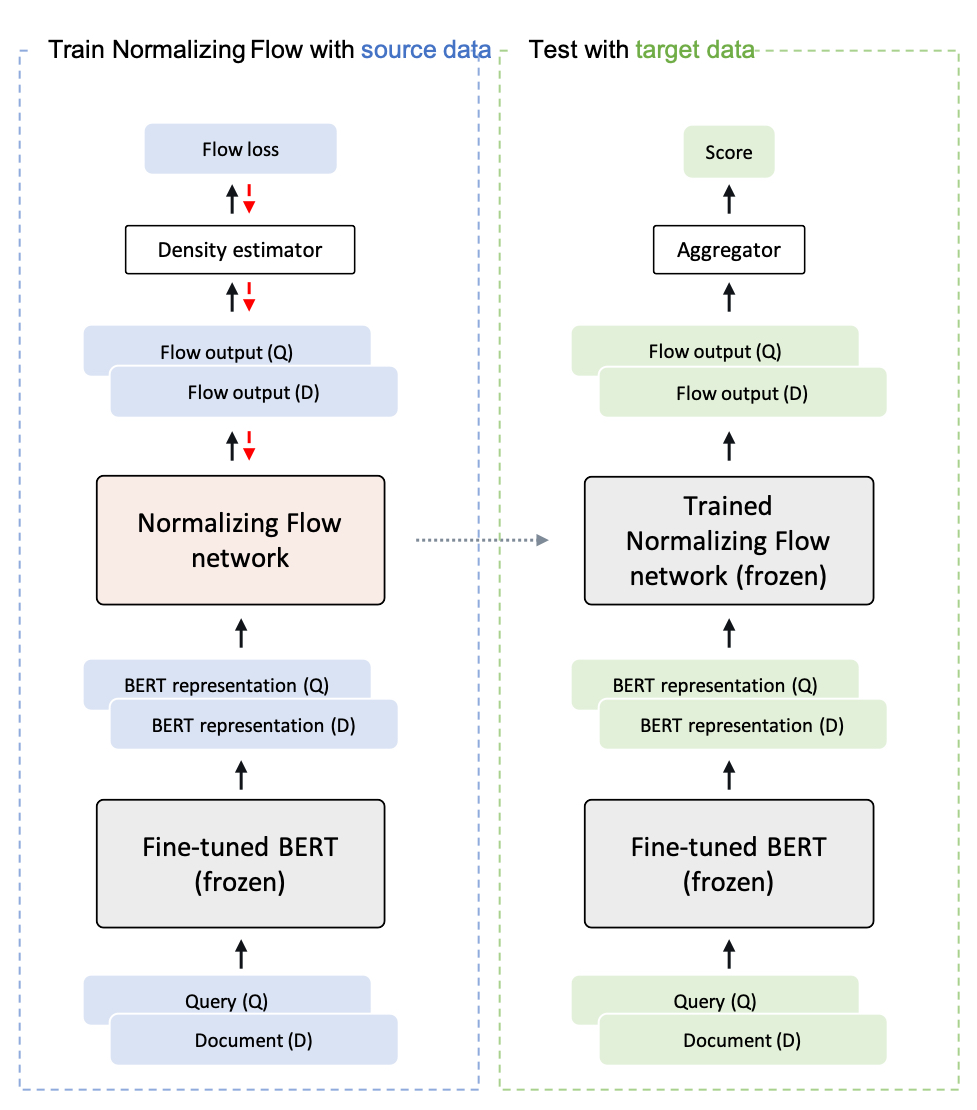}
    \caption{Normalizing Flow}
  \end{subfigure} \hfill
  \begin{subfigure}[b]{0.38\textwidth}
    \centering
    \includegraphics[width=\linewidth]{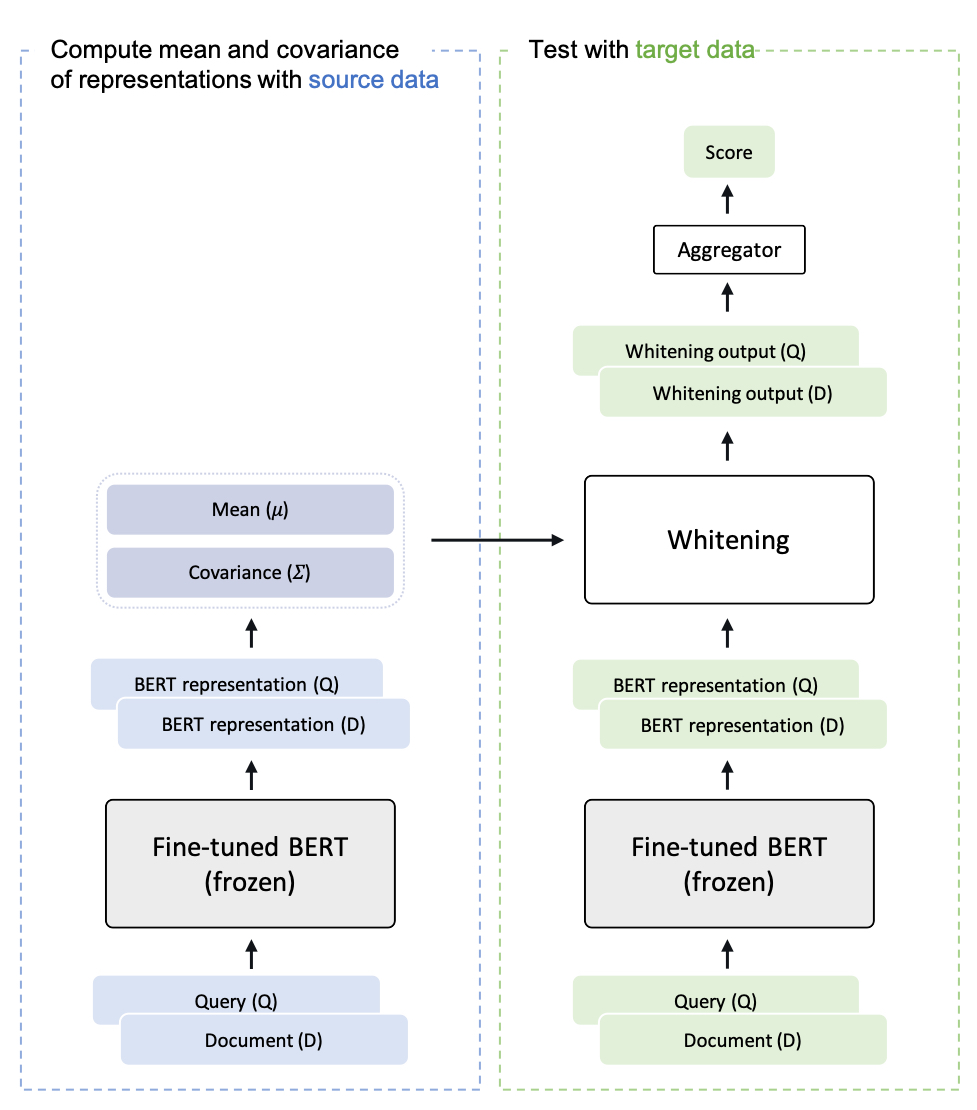}
    \caption{Whitening}
  \end{subfigure}
  \caption{Train and test procedures of Normalizing Flow and whitening. To apply a post-processing method to DR, we first fine-tune the BERT for DR as shown in (a). In (b), we train a Normalizing Flow network while keeping the fine-tuned BERT frozen. For whitening in (c), we pre-compute the mean and covariance using the fine-tuned BERT and apply the whitening filter during the test time. "Aggregator" in the test figures refers to the computation of the relevance score based on cosine similarity. For (b) and (c), the source and target data are the same for the ID re-ranking, and they are different for the OOD re-ranking.
  }
  \label{fig: methods}
  \vspace{-0.4cm}
\end{figure*}

\subsubsection{Normalizing Flow}
Normalizing Flow transforms a simple and tractable distribution into the target data distribution by applying a series of invertible and (almost everywhere) differentiable mappings \cite{tabak2010density,tabak2013family,kobyzev2020normalizing}. By doing this, flow-based models can easily infer the target density as in Eq.~(\ref{Inference in Flow}) and perform sampling from it as in Eq.~(\ref{Sampling in Flow}).
\begin{align}
    p_{X} (x) = p_{\text{Z}} (f(x)) \, | \text{det}\, \mathbf{D}f(x) |
    \label{Inference in Flow} \\
    x = g(z) \quad \text{where} \quad z \sim p_{Z}(z)
    \label{Sampling in Flow}
\end{align}
where $Z$ and $X$ are random variables of the simple and target distributions respectively, and $p_{X}$ and $p_{Z}$ are the densities of $X$ and $Z$, respectively. $f$ is a function that maps $X$ to $Z$, $g$ is the inverse of $f$, and $\mathbf{D}f(x) = \frac{\partial f}{\partial x}$ is the Jacobian matrix of $f$.

As illustrated in Figure~\ref{fig: methods}(b)-left, we aim to transform BERT representations of query and document into the standard Gaussian distribution using the Normalizing Flow's density estimation. To do so, in Eq. (\ref{Inference in Flow}), we set $p_{Z}$ as the density of the standard Gaussian distribution. Then, we maximize the likelihood of representations, that is, $p_{X}$ in Eq. (\ref{Inference in Flow}). Consequently, the flow function $f$ is trained to transform the representations $\mathbf{X}$ to follow the standard Gaussian distribution. Among the various flow-based models, we use NICE \cite{dinh2014nice} and Glow \cite{kingma2018glow}.

For DR, we propose two different implementations of Normalizing Flow. In \textit{token-wise} implementation, Normalizing Flow is applied to each \textit{token representation} of BERT. In \textit{sequence-wise} implementation, Normalizing Flow is applied to \textit{sequence representations} made by aggregating token representations.
Both implementations are applicable to single-vector models, such as RepBERT, but only the token-wise method is applicable to multi-vector models, such as ColBERT, because they do not utilize the sequence representation.

\subsubsection{Whitening}
Whitening is a linear transformation that renders the data distribution spherical. That is, it eliminates the structures of location, scale, and correlations in the distribution \cite{friedman1987exploratory}. Let $\{x_i\}_{i=1}^{N}$ be a set of BERT representations where each $x_i$ is a $D$-dimensional row vector. To apply whitening, a mean vector $\mu$ and an unbiased covariance matrix $\Sigma$ of $\{x_i\}_{i=1}^{N}$ are computed as $\mu = \frac{1}{N} \sum_{i=1}^{N} x_i$ and $\Sigma = \frac{1}{N-1} \sum_{i=1}^{N}(x_i - \mu)^{T}(x_i - \mu)$. 
By SVD and with some following calculations, the whitened representation vector $z_i$ can be presented as
\begin{align}
    z_i = (x_i - \mu) \; U \sqrt{\Lambda^{-1}},
    \label{Whitened representation}
\end{align}
where $\Lambda \in \mathbb{R}^{D \times D}$ is a diagonal matrix with positive diagonals, $U \in \mathbb{R}^{D \times D}$ is an orthogonal matrix, and each $z_i$ follows a distribution of zero mean and identity covariance~\cite{su2021whitening}. 

For DR, we first pre-compute the mean and covariance using a train set of the source data as shown in Figure~\ref{fig: methods}(c)-left. For the inference shown in Figure~\ref{fig: methods}(c)-right, we obtain the whitened output $z_i$ in Eq.~(\ref{Whitened representation}) using the pre-computed mean and covariance. The following aggregator then computes the cosine similarities between those $z_i$'s from queries and documents. As in the Normalizing Flow, we consider both token-wise and sequence-wise implementations of whitening.

\subsubsection{Metrics}
For a set of given representations, we measure the degree of isotropy by utilizing two metrics. Following Yu et al.~\cite{yu2022rare} and Wang et al.~\cite{wang2019improving}, we employ $I(\mathbf{W})$ suggested by Mu et al.~\cite{mu2017all} for measuring the isotropy of representation vectors.
The value of $I(\mathbf{W})$ ranges from 0 to 1, and representations having higher $I(\mathbf{W})$ tend to follow nearly isotropic Gaussian distribution.

Additionally, the average cosine similarity between representations, $avgcos$, is adopted for measuring isotropy following Ethayarajh et al.~\cite{ethayarajh2019contextual}. Representations vectors that are isotropically spread around the origin would have $avgcos$ values close to zero.
\vspace{-0.2cm}

\subsection{Robustness for Out-of-distribution Data}
Under the hypothesis that NRMs having representations that follow isotropic distribution are less likely to overfit to the characteristics of the training dataset, we compare OOD performances of DR models with and without post-processing methods. In each case, we train the DR model on the source data and evaluate the re-ranking performance on the different target data. As we consider three different datasets in this work, we investigate six different combinations of OOD experiments where the source data differs from the target data.
This concept of evaluating robustness is commonly addressed as OOD generalizability on an unforeseen corpus~\cite{wu2021neural}, and we will interchangeably use the terms OOD robustness and OOD generalizability in this work.
\section{Experiments}
\vspace{-0.2cm}

\subsection{Experimental Settings}
\subsubsection{Models, Datasets, and Ranking Metrics}
We investigate two DR models, ColBERT \cite{khattab2020colbert} and RepBERT \cite{zhan2020repbert}.
As for the datasets, we examined three popular document re-ranking datasets, Robust04~\cite{voorhees2004overview}, WebTrack 2009 (ClueWeb09b)~\cite{callan2009clueweb09}, and MS-MARCO~\cite{nguyen2016ms} as in \cite{macavaney2019cedr}. For Robust04, we used the document collections from TREC Disks 4 and 5\footnote{520K documents, 7.5K triplet data samples, https://trec.nist.gov/data-disks.html}. For ClueWeb09b, the document collections from ClueWeb09b\footnote{50M web pages, 4.5K triplet data samples, https://lemurproject.org/clueweb09/} of WebTrack 2009 were used. We also used the large document collections~\footnote{22G documents, 372K triplet data samples, https://microsoft.github.io/msmarco/TREC-Deep-Learning-2019} from MS-MARCO. To evaluate the ranking models, we used NDCG@10 and MRR@10 as the performance metrics.

\subsubsection{Training and Optimization}
Following Huston et al.~\cite{huston2014parameters}, we divided each of Robust04 and ClueWeb09b into five folds: three folds for training, one for validation, and the remaining one for the test. For MS-MARCO, we divided the dataset into training, validation, and test data.
Using the three performance values for three different random seeds, we conducted one-tailed $t$-test under the assumption of homoscedasticity.

For fine-tuning BERT, we used a learning rate of 1e-4, batch size of 16, and an Adam optimizer. Following the advice from \cite{zhang2020revisiting}, we set the maximum epoch to be 30 for Robust04 and ClueWeb09b and 10 for MS-MARCO. We then selected the model with the highest validation performance among the checkpoints recorded after each epoch during training.

To train Normalizing Flow, we stack a Normalizing Flow network at the top of the fine-tuned BERT and train it in an unsupervised manner for either ten epochs (Robust04 and ClueWeb09b) or three epochs (MS-MARCO) with a learning rate of 1e-4, while keeping the fine-tuned BERT weights frozen. For the NICE network, we used a five-layer network with 1000 units in each layer. For the Glow network, we used two levels and depth of three following \cite{li2020sentence}. For both NICE and Glow, we used a learning rate of 1e-4.

For whitening, we pre-compute mean and covariance. As different documents are selected in each epoch, we collect training data for ten epochs (Robust04 and ClueWeb09b) or three epochs (MS-MARCO) and use the collected representation vectors for the pre-computation of mean and covariance. For all the experiments, we tuned the hyper-parameters only lightly.

Our experiments are implemented with Python 3.8, Torch 1.9.0, Transformers 4.12.5, and Huggingface-hub 0.2.1. We fine-tuned the pre-trained BERT-base-uncased model provided by the huggingface transformers. We used RTX3090 GPUs, each of which has 25.6G memory.
\vspace{-0.2cm}

\subsection{Experimental Results}
\vspace{-0.1cm}

\subsubsection{Isotropic Representations Improve Re-ranking Performance}
In Table \ref{tab: colbert, performance}, we compare the re-ranking performance of fine-tuned ColBERT before and after applying Normalizing Flow or whitening to the BERT representations. Because ColBERT computes the cosine similarity between a query and a document's multiple token representations, we perform the token-wise post-processing.
The results demonstrate that the degree of isotropy is enhanced and the re-ranking performance is improved after the post-processing.
Across all three datasets, all of whitening, NICE, and Glow methods improve the re-ranking performance of the fine-tuned ColBERT by from 2.4\% to 8.1\% on NDCG@10. It can be confirmed that transforming BERT token representations to follow an isotropic distribution consistently improves the performance of ColBERT.

\begin{table}[t]
    \centering
    \caption{Re-ranking performance of ColBERT. We compare the performance of a fine-tuned model before and after post-processing BERT representations. NDCG at 10 (NDCG@10) and MRR at 10 (MRR@10) are used as re-ranking metrics, and $I(\mathbf{W})$ and $avgcos(\mathbf{W})$ are used to measure the isotropy of representations. Note: \normalfont{$^{*} p \leq 0.05$, $^{**} p \leq 0.01$ (1-tailed).}}
{\scriptsize
\begin{tabular}{c|l|ll|cc}
\hline
Dataset & Method & NDCG@10 & MRR@10 & $I(\mathbf{W})$ & $avgcos(\mathbf{W})$ \\
\hline
\multirow{4}{*}{MS-MARCO} & Fine-tuning (Ft) & 0.590 & 0.866 & 0.611 & 0.223 \\
& Ft  $\rightarrow$ Whitening & 0.630 (6.8\%)$^{*}$ & 0.901 (4.0\%)$^{*}$ & 0.918 & 0.009 \\
& Ft $\rightarrow$ NICE & 0.622 (5.4\%)$^{*}$ & 0.890 (2.8\%) & 0.636 & 0.191 \\
& Ft $\rightarrow$ Glow & \textbf{0.638 (8.1\%)}$^{*}$ & \textbf{0.914 (5.5\%)$^{*}$} & 0.837 & 0.027 \\
\hline

\multirow{4}{*}{Robust04} & Fine-tuning (Ft)  & 0.402 & 0.622 & 0.602 & 0.225 \\
& Ft $\rightarrow$ Whitening  & \textbf{0.423 (5.2\%)}$^{**}$ & \textbf{0.654 (5.2\%)$^{*}$} & 0.891 & 0.016 \\
& Ft $\rightarrow$ NICE & 0.412 (2.4\%)$^{*}$ & 0.643 (3.4\%)$^{**}$ & 0.769 & 0.064 \\
& Ft $\rightarrow$ Glow  & 0.412 (2.4\%)$^{*}$ & 0.637 (2.4\%)$^{*}$ & 0.824 & 0.031 \\
\hline

\multirow{4}{*}{ClueWeb09b} & Fine-tuning (Ft) & 0.288 & 0.514 & 0.614 & 0.219 \\
& Ft $\rightarrow$ Whitening & 0.301 (4.7\%)$^{**}$ & 0.521 (1.4\%) & 0.915 & 0.009 \\
& Ft $\rightarrow$ NICE & \textbf{0.305 (6.1\%)}$^{**}$ & \textbf{0.529 (3.0\%)} & 0.757 & 0.074 \\
& Ft $\rightarrow$ Glow & 0.300 (4.2\%)$^{**}$ & 0.522 (1.5\%) & 0.857 & 0.020 \\
\hline
\end{tabular}
}
    \label{tab: colbert, performance}
    \vspace{-0.3cm}
\end{table}

\begin{table}
    \centering
    \caption{Re-ranking performance of RepBERT. We compare the performance of a fine-tuned model before and after enforcing isotropy on BERT representations. For RepBERT, we compare the performances of token-wise and sequence-wise representation transformation methods. Note: \normalfont{$^{*} p \leq 0.05$, $^{**} p \leq 0.01$ (1-tailed).}}
{\tiny
\begin{tabular}{c|l|ll|ll}
\hline
\multirow{2}{*}{Dataset} & \multirow{2}{*}{Method} & \multicolumn{2}{c|}{Token-wise Method} & \multicolumn{2}{c}{Sequence-wise Method} \\
& & NDCG@10 & MRR@10 & NDCG@10 & MRR@10 \\
\hline
\multirow{4}{*}{MS-MARCO} & Fine-tuning (Ft) & 0.330 & 0.675 & 0.330 & 0.675 \\
& Ft  $\rightarrow$ Whitening & 0.388 (17.3\%)$^{**}$ & 0.724 (7.3\%)$^{*}$ & \textbf{0.406 (22.8\%)}$^{**}$ & 0.724 (7.3\%)$^{*}$ \\
& Ft $\rightarrow$ NICE & 0.339 (2.6\%) & 0.695 (2.9\%) & 0.337 (2.1\%) & 0.687 (1.9\%) \\
& Ft $\rightarrow$ Glow & 0.365 (10.4\%)$^{*}$ & 0.718 (6.4\%) & 0.398 (20.4\%)$^{**}$ & \textbf{0.728 (7.9\%)}$^{*}$ \\ \hline

\multirow{4}{*}{Robust04} & Fine-tuning (Ft) & 0.344 & 0.543 & 0.344 & 0.543 \\
& Ft $\rightarrow$ Whitening & \textbf{0.373 (8.5\%)}$^{*}$ & \textbf{0.601 (10.6\%)}$^{**}$ & 0.347 (0.8\%) & 0.575 (5.8\%)$^{*}$ \\
& Ft $\rightarrow$ NICE & 0.352 (2.4\%) & 0.572 (5.2\%)$^{*}$ & 0.333 (-3.1\%) & 0.535 (-1.5\%) \\
& Ft $\rightarrow$ Glow & 0.371 (7.8\%)$^{*}$ & 0.589 (8.3\%)$^{*}$ & 0.354 (2.9\%) & 0.566 (4.2\%) \\ \hline

\multirow{4}{*}{ClueWeb09b} & Fine-tuning (Ft) & 0.193 & 0.373 & 0.193 & 0.373 \\
& Ft  $\rightarrow$ Whitening & \textbf{0.237 (23.3\%)}$^{**}$ & \textbf{0.451 (20.9\%)}$^{**}$ & 0.219 (13.9\%)$^{*}$ & 0.429 (14.9\%)$^{**}$ \\
& Ft $\rightarrow$ NICE & 0.208 (8.0\%) & 0.403 (8.1\%) & 0.183 (-5.2\%) & 0.374 (0.1\%) \\
& Ft $\rightarrow$ Glow & 0.233 (21.0\%)$^{**}$ & 0.440 (17.9\%)$^{**}$ & 0.228 (18.2\%)$^{*}$ & 0.434 (16.4\%)$^{**}$ \\
\hline
\end{tabular}
}
    \label{tab: repbert, performance}
    \vspace{-0.3cm}
\end{table}

Table \ref{tab: repbert, performance} presents the results for RepBERT. We compare the performance improvement of both token-wise and sequence-wise post-processing methods.
The results in Table \ref{tab: repbert, performance} show that the performance is improved for all the cases except for when NICE is applied as a sequence-wise processing method for Robust04 or ClueWeb09b. Between token-wise and sequence-wise methods, it can be observed that token-wise performs better for Robust04 and ClueWeb09b and sequence-wise method performs better for MS-MARCO. We provide a discussion on this issue in Section~\ref{subsec:token_sequence}.
Among the three post-processing methods, whitening almost always achieves the best performance for each dataset as long as a token-wise method is used for Robust04 or ClueWeb09b and a sequence-wise method is used for MS-MARCO. Overall, whitening improves the performance of the fine-tuned model by 8.5\% $\sim$ 23.3\% on NDCG@10 over the three datasets.

\begin{table*}[t]
    \centering
    \caption{Out-of-distribution generalizability of ColBERT. Performance and isotropy are evaluated on the target data using the models trained with the source data. Note: \normalfont{$^{*} p \leq 0.05$, $^{**} p \leq 0.01$ (1-tailed).}}
{\scriptsize
\begin{tabular}{cc|l|ll|cc}
\hline
Source data & Target data & Method   & NDCG@10 & MRR@10 & $I(\mathbf{W})$ & $avgcos(\mathbf{W})$ \\
\hline \hline

\multirow{4}{*}{MS-MARCO} & \multirow{4}{*}{Robust04} & Fine-tuning (Ft) &0.343 & 0.561 & 0.588 & 0.248 \\
& & Ft  $\rightarrow$ Whitening & 0.412 (19.9\%)$^{**}$ & 0.644 (14.7\%)$^{**}$ & 0.849 & 0.022 \\
& & Ft $\rightarrow$ NICE & 0.412 (20.1\%)$^{**}$ & 0.632 (12.7\%)$^{**}$ & 0.677 & 0.138 \\
& & Ft $\rightarrow$ Glow & \textbf{0.429 (25.0\%)}$^{**}$ & \textbf{0.658 (17.2\%)}$^{**}$ & 0.799 & 0.041 \\ \hline

\multirow{4}{*}{MS-MARCO} & \multirow{4}{*}{ClueWeb09b} & Fine-tuning (Ft) & 0.296 & 0.535 & 0.546 & 0.323 \\
& & Ft  $\rightarrow$ Whitening & 0.310 (4.6\%) & 0.546 (2.1\%) & 0.894 & 0.012                 \\
& & Ft $\rightarrow$ NICE & \textbf{0.311 (5.0\%)} & \textbf{0.551 (3.0\%)}$^{**}$ & 0.656 & 0.165                    \\
& & Ft $\rightarrow$ Glow & 0.307 (3.7\%) & 0.551 (2.9\%)$^{*}$ & 0.835 & 0.028      \\ \hline \hline

\multirow{4}{*}{Robust04} & \multirow{4}{*}{MS-MARCO} & Fine-tuning (Ft) & 0.383 & 0.666 & 0.627 & 0.206 \\
& & Ft  $\rightarrow$ Whitening &  0.417 (8.9\%)$^{**}$ & 0.708 (6.2\%)$^{*}$ & 0.809 & 0.037                   \\
& & Ft $\rightarrow$ NICE & 0.419 (9.2\%)$^{**}$ & 0.710 (6.6\%)$^{*}$ &0.761 & 0.065                    \\
& & Ft $\rightarrow$ Glow & \textbf{0.419 (9.4\%)}$^{**}$ & \textbf{0.716 (7.6\%)}$^{**}$ & 0.786 & 0.048       \\ \hline

\multirow{4}{*}{Robust04} & \multirow{4}{*}{ClueWeb09b} & Fine-tuning (Ft) & 0.219 & 0.408 & 0.618 & 0.219 \\
& & Ft  $\rightarrow$ Whitening & 0.242 (10.2\%)$^{**}$ & 0.445 (9.0\%)$^{**}$ & 0.787 & 0.047                   \\
& & Ft $\rightarrow$ NICE & \textbf{0.245 (11.4\%)}$^{**}$ & \textbf{0.453 (11.0\%)}$^{**}$ & 0.746 & 0.074                     \\
& & Ft $\rightarrow$ Glow & 0.243 (10.6\%)$^{**}$ & 0.447 (9.5\%)$^{**}$ & 0.769 & 0.057    \\ \hline \hline

\multirow{4}{*}{ClueWeb09b} & \multirow{4}{*}{MS-MARCO} & Fine-tuning (Ft) & 0.451 & 0.795 & 0.611 & 0.232 \\
& & Ft  $\rightarrow$ Whitening  & 0.469 (4.1\%)$^{*}$ & 0.804 (1.5\%) & 0.901 & 0.011                   \\
& & Ft $\rightarrow$ NICE  & \textbf{0.476 (5.7\%)}$^{**}$ & \textbf{0.820 (3.2\%)}$^{*}$ & 0.765 & 0.069                     \\
& & Ft $\rightarrow$ Glow  & 0.470 (4.3\%)$^{*}$ & 0.814 (2.4\%)$^{*}$ & 0.852 & 0.021      \\ \hline

\multirow{4}{*}{ClueWeb09b} & \multirow{4}{*}{Robust04} & Fine-tuning (Ft) & 0.359 & 0.571 & 0.606 & 0.237   \\
& & Ft  $\rightarrow$ Whitening  & 0.402 (12.1\%)$^{**}$ & \textbf{0.623 (9.1\%)}$^{**}$ & 0.842 & 0.024                   \\
& & Ft $\rightarrow$ NICE  & 0.398 (10.8\%)$^{**}$ & 0.610 (6.7\%)$^{*}$ & 0.743 & 0.080                    \\
& & Ft $\rightarrow$ Glow  & \textbf{0.405 (12.8\%)}$^{**}$ & 0.620 (8.4\%)$^{**}$ & 0.800 & 0.041      \\
\hline
\end{tabular}
}
    \label{tab: colbert, domain generalization performance}
    \vspace{-0.3cm}
\end{table*}

\subsubsection{Isotropic Representations Make DR Models Robust to OOD Data.}
\begin{figure*}[t]
  \setcounter{figure}{3}
  \centering
  \hspace*{\fill}%
  \subcaptionbox{Pre-training}{\includegraphics[width=0.24\linewidth]{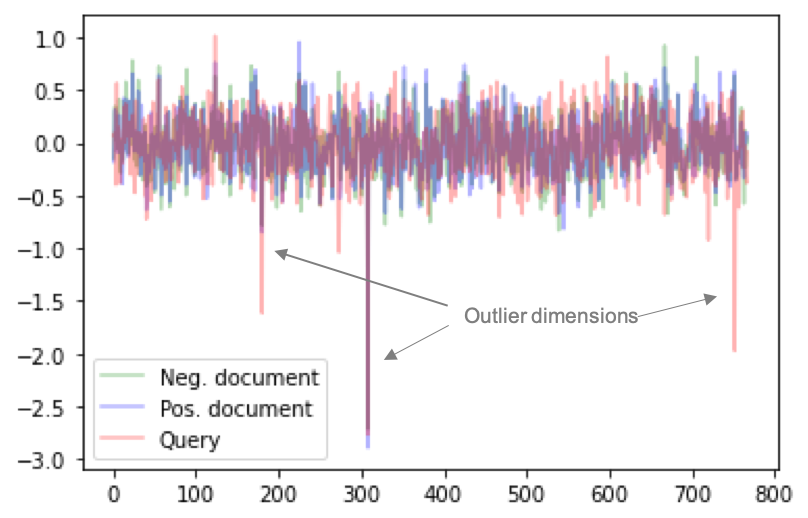}}
  \hfill
  \subcaptionbox{Fine-tuning (Ft)}{\includegraphics[width=0.24\linewidth]{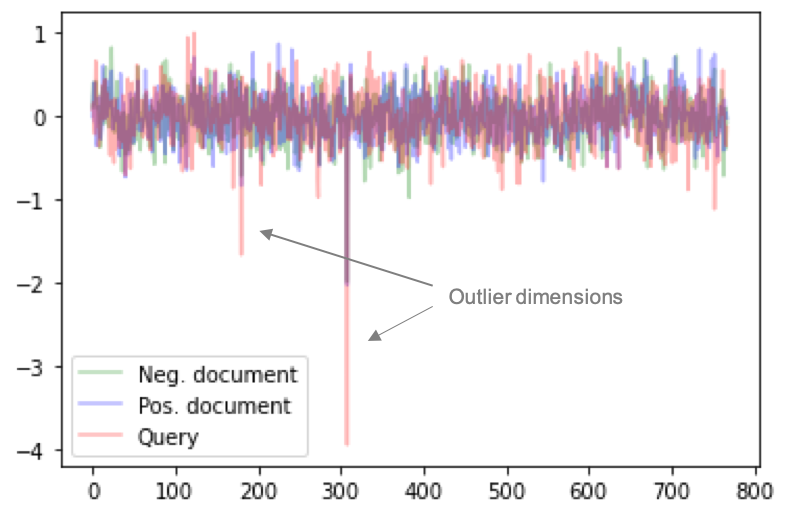}}
  \hfill
  \subcaptionbox{Ft $\rightarrow$ Glow}{\includegraphics[width=0.24\linewidth]{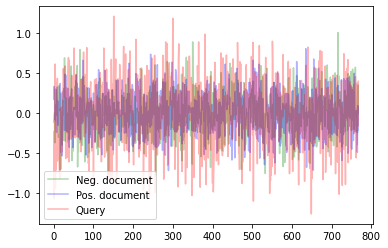}}
  \hfill
  \subcaptionbox{Ft $\rightarrow$ Whitening}{\includegraphics[width=0.24\linewidth]{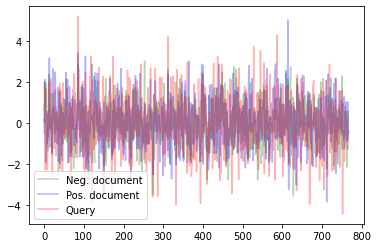}}
  \hspace*{\fill}%
  \vspace{-0.2cm}
  \caption{Visualization of sequence representation vectors of RepBERT:  representations of a randomly sampled triplet of a query, positive document, and negative document from Robust04 are shown. The outlier dimensions with spiky values shown in representations of pre-trained BERT (a) and fine-tuned BERT (b) are tempered by enforcing isotropy as shown in (c) and (d).}
  \label{fig: rep dim analysis}
  \vspace{-0.4cm}
\end{figure*}

We show the effectiveness of isotropic representations for OOD generalizability in Table \ref{tab: colbert, domain generalization performance}.
In this result for ColBERT, it can be observed that the baseline OOD generalizability of a fine-tuned model is always improved by applying a post-processing method that enforces isotropy to the BERT representations. 
Performance of NDCG@10 is improved by 5.0\% $\sim$ 25.0\%, and MRR@10 by 3.0\% $\sim$ 17.2\%, across the three datasets.
The overall result indicates that isotropically distributed BERT representations are robust in the sense that they can perform better for an OOD dataset that has been unseen during the training.
 
In particular, when ColBERT is trained with MS-MARCO as the source data and tested with the target data of either Robust04 or ClueWeb09b, the OOD performance with isotropy enhancement even surpasses the ID performance. For example, an ID performance on NDCG@10 of ColBERT fine-tuned on Robust04 is 0.402, and the OOD performance when the source data is MS-MARCO is 0.343. However, when we post-process the model using Glow, the performance is improved to 0.429, which is even higher than the baseline ID performance.
This result can be interpreted to be surprising because the OOD models trained \textit{without} the target data outperformed the ID models that were trained \textit{with} the target data.
Although we did not include the results for RepBERT, similar results can be obtained for RepBERT as well. We present the OOD generalizability result for RepBERT in the supplementary material.
\section{Discussion}
\vspace{-0.2cm}

\subsection{Handling of Outlier Dimensions}
The existence of outlier dimensions with extreme values has been known for BERT and other language models' representations~\cite{luo2020positional,kovaleva2021bert,timkey2021all}. For example, Luo et al.~\cite{luo2020positional} pointed out that the outlier dimensions can have a negative impact on task performance and showed that an improvement was possible by simply clipping the outliers.
In this study, we show that the outlier dimensions can also be observed in BERT representations fine-tuned for DR as shown in Figure~\ref{fig: rep dim analysis}(b). Even though we didn't explicitly aim to handle such outlier dimensions, it can be seen in Figure~\ref{fig: rep dim analysis}(c) and \ref{fig: rep dim analysis}(d) that the outlier dimensions are tempered and become hardly observable after enforcing isotropy.

\subsection{Token-wise vs. Sequence-wise Transformation}
\label{subsec:token_sequence}
As explained in Section~\ref{subsection:enforcing_isotropy}, isotropy can be enforced with either token-wise or sequence-wise transformation. For RepBERT, which is a single-vector DR model, both methods can be applied. The results for RepBERT in Table~\ref{tab: repbert, performance} show that the token-wise method outperforms the sequence-wise method on Robust04 and ClueWeb09b, and vice versa on MS-MARCO. A possible explanation for the results is the length of the queries. While MS-MARCO tends to have long queries of complete sentences, Robust04 and ClueWeb09b tend to have keyword-based short queries \cite{jung2022semi}. Because the representations of the queries play an important role in NRMs, it might be natural for token-wise transformation to perform well on Robust04 and ClueWeb09b having short queries.

\subsection{Robustness and OOD Generalizability}
As explained in Section~\ref{subsec:robustness}, robustness is essential for DR, and OOD generalizability can be considered as one of the most important factors of robustness. From Table~\ref{tab: colbert, domain generalization performance}, it can be confirmed that a significant improvement in OOD generalizability can be attained by enforcing isotropy. When the source data is MS-MARCO, we were able to make the OOD performance with isotropy enhancement even surpass the baseline ID performance. While we have analyzed only the most fundamental scenarios, the results imply that it might be possible to further improve the robustness of DR models by enforcing isotropy and concurrently considering more complex schemes such as the use of multiple source datasets, advancement of the methods for enforcing isotropy, etc.
\section{Conclusion}
\vspace{-0.2cm}

In this work, we have confirmed that the representations of BERT-based DR models are anisotropically distributed. Such an anisotropy can negatively affect the performance of DR models. We applied post-processing methods such as Normalizing Flow and whitening and have shown how to apply the methods in token-wise and sequence-wise manners in DR. With the proposed methods, we were able to improve the re-ranking performance of ColBERT and RepBERT for all the cases that we have studied. In addition to the commonly studied re-ranking with an in-distribution dataset, we have shown that isotropy of representations can be an essential factor for enhancing robustness of DR models. For the out-of-distribution tasks, we were able to achieve large improvements in many cases. Based on our results, isotropy can be deemed to be a crucial element for studying and improving the representations of DR models.

\subsubsection{Acknowledgements}
This work was supported by Naver corporation (Development of an Improved Neural Ranking Model.), National Research Foundation of Korea (NRF) grant funded by the Korea government (MSIT) (No. NRF-2020R1A2C2007139), and IITP grant funded by the Korea government (No. 2021-0-01343, Artificial Intelligence Graduate School Program (Seoul National University)).

\bibliographystyle{splncs04}
% \bibliographystyle{MathPhySci}
% {\tiny
\bibliography{main.bbl}




% LNCS Homepage, \url{http://www.springer.com/lncs}. Last accessed 4
% \end{thebibliography}

\section{Appendix}
As supplementary material, we provide the OOD generalizability results of RepBERT. Similar to the results of ColBERT, isotropy enhancement improves the OOD performance of RepBERT. Across all six combinations of the source and target data, post-processing improved the re-ranking performance of RepBERT by from 6.6\% to 39.4\% on NDCG@10. We can also find out that the OOD performance after post-processing outperforms the baseline ID performance in some cases. For example, the ID performance (NDCG@10) of the fine-tuned RepBERT on ClueWeb09b is 0.193. When we change the source data into MS-MARCO, the performance drops to 0.159. However, if we apply token-wise whitening, the OOD performance is improved to 0.209, which is even higher than the ID performance.

\begin{table*}[]
    \centering
    \caption{Out-of-distribution generalizability of RepBERT. Performance and isotropy are evaluated on the target data using the models trained with the source data. Note: \normalfont{$^{*} p \leq 0.05$, $^{**} p \leq 0.01$ (1-tailed).}}
{\tiny
\begin{tabular}{cc|c|ll|ll}
\hline
\multirow{2}{*}{Source data} & \multirow{2}{*}{Target data} & \multirow{2}{*}{Method} & \multicolumn{2}{c|}{Token-wise Method} & \multicolumn{2}{c}{Sequence-wise Method} \\
& & & NDCG@10 & MRR@10 & NDCG@10 & MRR@10 \\
\hline \hline

\multirow{4}{*}{MS-MARCO} & \multirow{4}{*}{Robust04} & Fine-tuning (Ft) & 0.216 & 0.389 & 0.216 & 0.389 \\
& & Ft  $\rightarrow$ Whitening & 0.280 (9.3\%)$^{**}$ & 0.469 (20.5\%)$^{**}$ & 0.300 (28.9\%)$^{**}$ & \textbf{0.494 (27.0\%)}$^{**}$ \\
& & Ft $\rightarrow$ NICE & 0.229 (6.1\%) & 0.421 (8.4\%) & 0.230 (6.5\%) & 0.411 (5.7\%) \\
& & Ft $\rightarrow$ Glow & 0.275 (27.3\%)$^{*}$ & 0.460 (18.4\%)$^{*}$ & \textbf{0.301 (39.4\%)}$^{**}$ & 0.489 (25.7\%)$^{**}$ \\ \hline

\multirow{4}{*}{MS-MARCO} & \multirow{4}{*}{ClueWeb09b} & Fine-tuning (Ft) & 0.159 & 0.328 & 0.159 & 0.328 \\
& & Ft  $\rightarrow$ Whitening & \textbf{0.209 (31.8\%)} & 0.399 (21.7\%)$^{*}$ & 0.207 (30.2\%) & 0.414 (26.3\%)$^{*}$ \\
& & Ft $\rightarrow$ NICE & 0.175 (10.2\%) & 0.356 (8.4\%) & 0.172 (8.4\%) & 0.349 (6.6\%) \\
& & Ft $\rightarrow$ Glow & 0.202 (27.3\%) & 0.408 (24.5\%)$^{*}$ & 0.209 (31.6\%) & \textbf{0.415 (26.6\%)}$^{*}$ \\ \hline \hline

\multirow{4}{*}{Robust04} & \multirow{4}{*}{MS-MARCO} & Fine-tuning (Ft) & 0.331 & 0.604 & 0.331 & 0.604 \\
& & Ft  $\rightarrow$ Whitening & \textbf{0.353 (6.6\%)}$^{*}$ & \textbf{0.628 (4.0\%)} & 0.352 (6.4\%) & 0.616 (1.9\%) \\
& & Ft $\rightarrow$ NICE & 0.335 (1.1\%) & 0.597 (-1.2\%) & 0.324 (-2.1\%) & 0.563 (-6.8\%) \\
& & Ft $\rightarrow$ Glow & 0.344 (3.7\%) & 0.603 (-0.3\%) & 0.343 (3.7\%) & 0.597 (-1.3\%) \\ \hline

\multirow{4}{*}{Robust04} & \multirow{4}{*}{ClueWeb09b} & Fine-tuning (Ft) & 0.162 & 0.338 & 0.162 & 0.338 \\
& & Ft  $\rightarrow$ Whitening & \textbf{0.202 (24.9\%)}$^{**}$ & 0.383 (13.0\%)$^{*}$ & 0.183 (13.2\%) & 0.380 (12.1\%) \\
& & Ft $\rightarrow$ NICE & 0.192 (18.8\%)$^{*}$ & 0.384 (13.5\%)$^{*}$ & 0.176 (8.8\%) & 0.376 (11.1\%) \\
& & Ft $\rightarrow$ Glow & 0.199 (23.0\%)$^{*}$ & \textbf{0.285 (13.6\%)}$^{*}$ & 0.187 (15.5\%)$^{*}$ & 0.379 (11.9\%) \\ \hline \hline

\multirow{4}{*}{ClueWeb09b} & \multirow{4}{*}{MS-MARCO} & Fine-tuning (Ft) & 0.347 & 0.632 & 0.347 & 0.632 \\
& & Ft  $\rightarrow$ Whitening & \textbf{0.379 (9.4\%)}$^{*}$ & \textbf{0.697 (10.3\%)}$^{*}$ & 0.372 (7.3\%) & 0.677 (7.1\%) \\
& & Ft $\rightarrow$ NICE & 0.352 (1.4\%) & 0.655 (3.7\%) & 0.338 (-2.5\%) & 0.631 (-0.1\%) \\
& & Ft $\rightarrow$ Glow & 0.375 (8.1\%) & 0.681 (7.7\%)$^{*}$ & 0.365 (5.4\%) & 0.663 (5.0\%) \\ \hline

\multirow{4}{*}{ClueWeb09b} & \multirow{4}{*}{Robust04} & Fine-tuning (Ft) & 0.289 & 0.486 & 0.289 & 0.486 \\
& & Ft  $\rightarrow$ Whitening & 0.328 (13.4\%)$^{**}$ & 0.534 (10.0\%)$^{*}$ & 0.309 (7.1\%)$^{*}$ & 0.512 (5.4\%) \\
& & Ft $\rightarrow$ NICE & 0.301 (4.4\%) & 0.504 (3.7\%) & 0.284 (-1.6\%) & 0.474 (-2.3\%) \\
& & Ft $\rightarrow$ Glow & \textbf{0.333 (15.3\%)}$^{**}$ & \textbf{0.541 (11.3\%)}$^{*}$ & 0.325 (12.5\%)$^{**}$ & 0.526 (8.3\%)$^{*}$ \\ \hline

\hline
\end{tabular}
}
    \label{tab: colbert, domain generalization performance}
\end{table*}
\end{document}